\documentclass{article}

\usepackage{microtype}
\usepackage{graphicx}
\usepackage{subfigure}
\usepackage{booktabs} 

\usepackage{hyperref}

\usepackage[accepted]{icml2024}

\usepackage{mathtools,amssymb,amsthm,bm}
\usepackage[Symbol]{upgreek}
\usepackage{pifont}
\newcommand{\cmark}{\ding{51}}%
\newcommand{\xmark}{\ding{55}}%

\newcommand\Nd{\mathbb{N}}
\newcommand\du{\mathrm{d}}
\newcommand\eu{\mathrm{e}}
\newcommand\Tu{\mathrm{T}}
\newcommand\Rd{\mathbb{R}}
\newcommand\Ed{\mathbb{E}}
\newcommand\Pd{\mathbb{P}}

\newcommand\Xc{\mathcal{X}}

\newcommand\ELBO{\operatorname{ELBO}}
\newcommand\KL{\operatorname{KL}}
\newcommand\rbr[1]{\left(#1\right)}
\newcommand\sbr[1]{\left[#1\right]}
\newcommand\cbr[1]{\left\{#1\right\}}

\newcommand\abs[1]{\left|#1\right|}
\newcommand\pderiv[2]{\frac{\partial #1}{\partial #2}}

\usepackage[capitalize,noabbrev]{cleveref}

\theoremstyle{plain}

\theoremstyle{definition}

\theoremstyle{remark}

\usepackage[textsize=tiny]{todonotes}

\icmltitlerunning{A Differentiable POGLM with Forward-Backward Message Passing}

\begin{document}
\twocolumn[
\icmltitle{A Differentiable Partially Observable Generalized Linear Model with Forward-Backward Message Passing}



\icmlsetsymbol{equal}{*}

\begin{icmlauthorlist}
\icmlauthor{Chengrui Li}{cse}
\icmlauthor{Weihan Li}{cse}
\icmlauthor{Yule Wang}{cse}
\icmlauthor{Anqi Wu}{cse}

\end{icmlauthorlist}

\icmlaffiliation{cse}{School of Computational Science \& Engineering, Georgia Institute of Technology, Atlanta, USA}

\icmlcorrespondingauthor{Chengrui Li}{cnlichengrui@gatech.edu}
\icmlcorrespondingauthor{Anqi Wu}{anqiwu@gatech.edu}

\icmlkeywords{Machine Learning, ICML}

\vskip 0.3in
]



\printAffiliationsAndNotice{}  

\begin{abstract}
The partially observable generalized linear model (POGLM) is a powerful tool for understanding neural connectivity under the assumption of existing hidden neurons. With spike trains only recorded from visible neurons, existing works use variational inference to learn POGLM meanwhile presenting the difficulty of learning this latent variable model. There are two main issues: (1) the sampled Poisson hidden spike count hinders the use of the pathwise gradient estimator in VI; and (2) the existing design of the variational model is neither expressive nor time-efficient, which further affects the performance. For (1), we propose a new differentiable POGLM, which enables the pathwise gradient estimator, better than the score function gradient estimator used in existing works. For (2), we propose the forward-backward message-passing sampling scheme for the variational model. Comprehensive experiments show that our differentiable POGLMs with our forward-backward message passing produce a better performance on one synthetic and two real-world datasets. Furthermore, our new method yields more interpretable parameters, underscoring its significance in neuroscience.
\end{abstract}

\section{Introduction}\label{sec:1}
Understanding neural connectivity is a critical research question in neuroscience. The generalized linear model (GLM) \citep{pillow2008spatio} with its variants \citep{linderman2016bayesian,roudi2015multi,li2024one} form a mainstream set of tools for inferring connectivity from neural populations. However, a nonnegligible problem in neural recording is that the recorded neurons are only a small part of the entire population in a certain target region of our interests. A GLM for such an incomplete problem is referred to as a partially observable GLM (POGLM) \citep{pillow2007neural,jimenez2014stochastic,linderman2017bayesian}, which considers both visible and hidden neurons.

The general goal of POGLM is to learn the model parameter set $\theta$, especially the connectivity between both visible and hidden neurons given only the spike trains $\bm X$ from visible neurons, and the spike trains from hidden neurons $\bm Z$ is the latent variable in POGLM. Variational inference (VI) \citep{blei2017variational} is the most commonly used method for solving such a latent variable model. In VI, the target is to maximize the observable data's evidence lower bound
\vspace{-0.03in}
\begin{equation}\label{eq:ELBO}
    \begin{split}
        & \ELBO(\bm X;\theta,\phi) \\
        = & \Ed_{q(\bm Z|\bm X;\phi)} \sbr{\ln p(\bm X,\bm Z;\theta) - \ln q(\bm Z|\bm X;\phi)} \\
        = & \ln p(\bm X;\theta) - \KL(q(\bm Z|\bm X;\phi)\|p(\bm Z|\bm X;\theta)) \\
        \leqslant & \ln p(\bm X;\theta)
    \end{split}
\end{equation}
w.r.t. $\theta$ and $\phi$, where $p(\bm X,\bm Z;\theta)$ is the generative model and $q(\bm Z|\bm X;\phi)$ is the variational model parameterized by $\phi$ approximating the posterior $p(\bm Z|\bm X;\theta)$. Maximizing Eq.~\ref{eq:ELBO} requires sampling hidden spike train $\bm Z$ from $q(\bm Z|\bm X;\phi)$ and computing the gradient estimator of $\ELBO(\bm X;\theta,\phi)$ w.r.t. $\theta$ and $\phi$.

However, results from existing works demonstrated the difficulties of solving such a complicated model.
Particularly, two issues have caught our attention:\\
(1) A good way of deriving $\pderiv{\ELBO(\bm X;\theta,\phi)}{\phi}$ is via the pathwise gradient estimator \cite{kingma2013auto}, which will be hindered with a discrete $\bm Z$. The only alternative is the score function gradient estimator which usually has higher variance than the pathwise gradient estimator \citep{paisley2012variational,bengio2013estimating,schulman2015gradient}.\\
(2) The sampling scheme of the variational model $\bm Z\sim q(\bm Z|\bm X;\phi)$ in most of the existing works is a GLM on hidden neurons \citep{jimenez2014stochastic,kajino2021differentiable,li2024forward}, which is only conditioned on history visible and sampled history hidden spikes. This design makes the sampling and inference procedure slow and omits the conditioning of hidden spikes on future visible spikes.

Given these issues, our paper aims to solve these two limitations in the existing works and study the POGLM more comprehensively. In Sec.~\ref{sec:2}, we will propose a new differentiable POGLM that enables the pathwise gradient estimator for VI. We will also introduce different variational distribution families for sampling hidden spikes, especially our newly proposed forward-backward message passing. In Sec.~\ref{sec:3}, we will conduct extensive comparisons between combinations of different inference methods (including the original POGLM, our newly proposed differentiable POGLM, and other intermediate models) $\times$ different variational sampling schemes on a synthetic dataset and two real-world neural datasets. The results will demonstrate the superiorities of our differentiable POGLM and our forward-backward message-passing sampling scheme.

\vspace{-0.1in}
\section{Models}\label{sec:2}
\vspace{-0.05in}
\subsection{Background: POGLM}
\begin{figure}
    \centering
    \includegraphics[width=\linewidth]{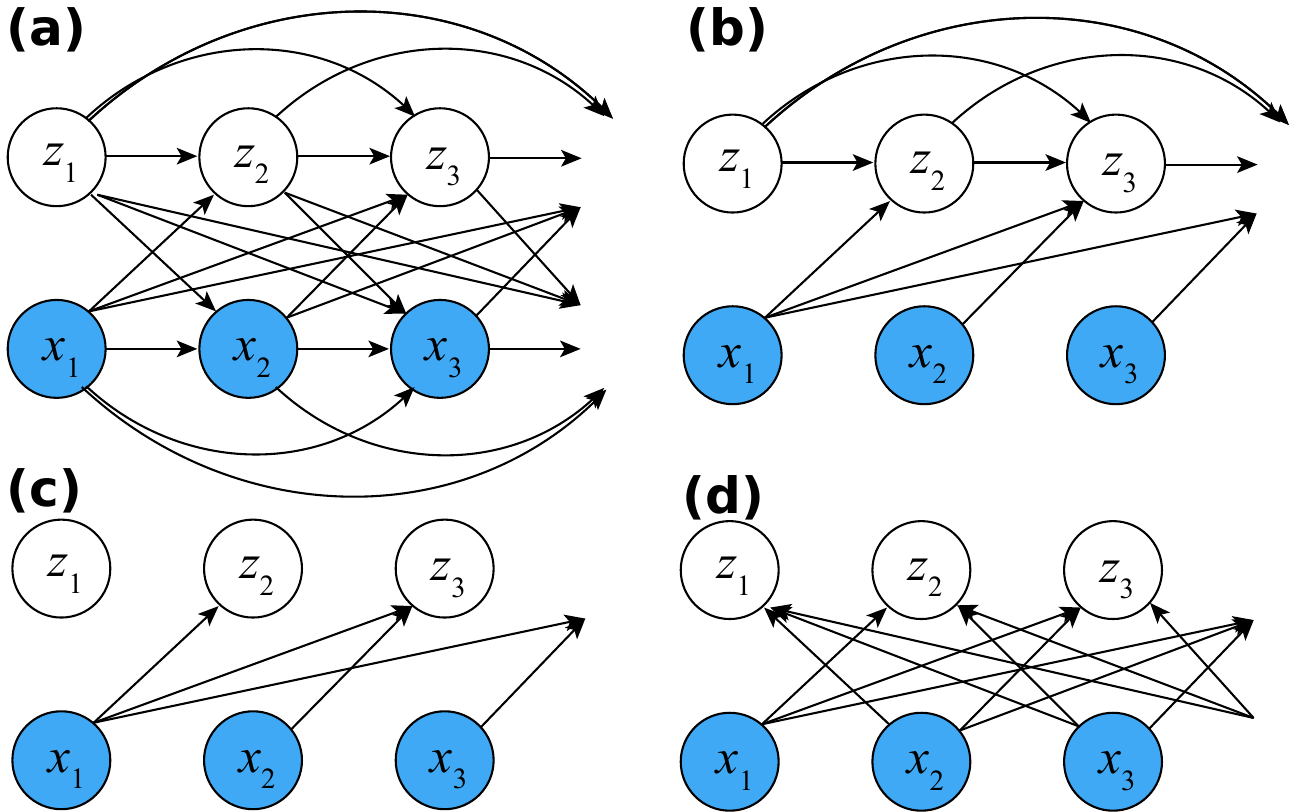}
    \vspace{-0.2in}
    \caption{\textbf{(a)}: The generative model of the complete POGLM $p(\bm X,\bm Z;\theta)$. \textbf{(b)}, \textbf{(c)}, \textbf{(d)}: The forward-self, forward, and forward-backward sampling scheme of the variational model $q(\bm Z|\bm X;\phi)$.}
    \label{fig:graphical_model}
    \vspace{-0.1in}
\end{figure}

\vspace{-0.05in}
\paragraph{Generative model.} We start from the partially observable generalized linear model (POGLM) \citep{pillow2007neural,jimenez2014stochastic,linderman2017bayesian} that studies the mutual interactions between neurons underlying the corresponding neural spike trains. Assume $V$ of $N$ neurons are visible and the remaining $H=N-V$ neurons are hidden. We denote $\bm X\in\Nd^{T\times V}$ as the observed spike train recorded from $V$ visible neurons across $T$ time bins, and $x_{t,v}$ as the spike count generated by the $v$-th visible neuron in the $t$-th time bin. $\bm Z\in\Nd^{T\times H}$ as the latent spike train recorded from $H$ hidden neurons across $T$ time bins, and $z_{t,h}$ as the spike count generated by the $h$-th hidden neuron in the $t$-th time bin. The complete generative model $p(\bm X,\bm Z;\theta)$ is depicted in Fig.~\ref{fig:graphical_model}(a) \citep{pillow2008spatio}. For a visible neuron $v$, its firing rate at time $t$ is
\vspace{-0.1in}
\begin{equation}\label{eq:GLM_v}
    \begin{split}
        f_{t,v} = & \sigma\Bigg(b_v + \sum_{v'=1}^V w_{v\gets v'} \cdot \rbr{\sum_{l=1}^L x_{t-l,v'}\ \psi_l}\\
        & + \sum_{h'=1}^H w_{v\gets h'} \cdot \rbr{\sum_{l=1}^L z_{t-l,h'}\ \psi_l}\Bigg),
    \end{split}
\end{equation}
and its spike count is generated by
\vspace{-0.1in}
\begin{equation}\label{eq:Poisson_v}
    x_{t,v}\sim\operatorname{Poisson}(f_{t,v}).
\vspace{-0.1in}
\end{equation}
$\sigma(\cdot)$ is a non-linear function (e.g., Softplus); $\bm b_V = [b_1, b_2, \dots, b_V]^\Tu\in\Rd^V$ is the background intensity vector of the $V$ neurons; $\bm W_{V\gets V} = \sbr{w_{v\gets v'}}_{V\times V}\in \Rd^{V\times V}$ is the weight matrix representing the weights from visible neurons to visible neurons; $\bm W_{V\gets H} = \sbr{w_{v\gets h'}}_{V\times H}\in \Rd^{V\times H}$ is the weight matrix representing the weights from hidden neurons to visible neurons; $\bm\psi = [\psi_1,\psi_2,\dots,\psi_L]^\Tu \in \Rd_+^L$ is the pre-defined basis function summarizing history spikes from $t-L$ to $t-1$. Similarly, for a hidden neuron $h$, its firing rate at time $t$ is
\vspace{-0.1in}
\begin{equation}\label{eq:GLM_h}
    \begin{split}
        f_{t,h} = & \sigma\Bigg(b_h + \sum_{v'=1}^V w_{h\gets v'} \cdot \rbr{\sum_{l=1}^L x_{t-l,v'}\ \psi_l}\\
        & + \sum_{h'=1}^H w_{h\gets h'} \cdot \rbr{\sum_{l=1}^L z_{t-l,h'}\ \psi_l}\Bigg),
    \end{split}
\end{equation}
and its spike count is generated by
\vspace{-0.1in}
\begin{equation}\label{eq:Poisson_h}
    z_{t,h} \sim \operatorname{Poisson}(f_{t,h})
\end{equation}
with parameters $\bm b_H = [b_1,b_2,\dots, b_H]^\Tu\in\Rd^H$; $\bm W_{H\gets V} = [w_{h\gets v'}]_{H\times V}\in \Rd^{H\times V}$; $\bm W_{H\gets H} = [w_{h\gets h'}]_{H\times H}\in\Rd^{H\times H}$.

Therefore, POGLM is a latent variable model whose learnable parameter set is $\theta = \cbr{\bm b,\bm W}$ where $\bm b$ and $\bm W$ can be presented in the form of block (partitioned) matrix/vector:
\vspace{-0.1in}
\begin{equation}
    \bm b = \begin{bmatrix}
        \bm b_V \\ \bm b_H
    \end{bmatrix} \in \Rd^N,\ \ 
    \bm W = \begin{bmatrix}
        \bm W_{V\gets V} & \bm W_{V\gets H} \\
        \bm W_{H\gets V} & \bm W_{H\gets H}
    \end{bmatrix} \in \Rd^{N\times N}.
\vspace{-0.1in}
\end{equation}

\paragraph{Variational inference.} Since POGLM is a latent variable model, the goal is to learn the model parameter $\theta$ while also inferring the latent variable $\bm Z$. Given the complicated nature of POGLM (Fig.~\ref{fig:graphical_model}(a)), there is no closed form posterior distribution $p(\bm Z|\bm X;\theta)$. Therefore, we need to choose a good variational model $q(\bm Z|\bm X;\phi)$ parameterized by $\phi$ to do variational inference (VI) \citep{blei2017variational}. We will discuss different choices of the variational models in Sec.~\ref{sec:sampling_schemes}. Now, we use a simple homogeneous Poisson variational model for illustration. The firing rate of a hidden neuron at time $t$ is given by
\vspace{-0.1in}
\begin{equation}\label{eq:homo_Poisson}
    f_{t,h} = \sigma(c_h),
\end{equation}
and the spike count is $z_{t,h} \sim \operatorname{Poisson}(f_{t,h})$. The variational parameter set is $\phi = \cbr{\bm c_H}$, where $\bm c_H = [c_1,\dots, c_H]^\Tu$.

With a selected variational model $q(\bm Z|\bm X;\phi)$, VI can be adopted. We maximize the model's evidence lower bound $\ELBO(\bm X;\theta,\phi)$ (Eq.~\ref{eq:ELBO}) w.r.t. $\theta$ and $\phi$, so that the learned $\theta$ could estimate the model's true parameter $\theta^{\text{true}}$ and the variational distribution $q(\bm Z|\bm X;\phi)$ could approximate the unknown posterior distribution $p(\bm Z|\bm X;\theta)$ of the latent variable $\bm Z$. Given the complicated form of the POGLM model (Fig.~\ref{fig:graphical_model}(a)), there is still no closed form of ELBO (Eq.~\ref{eq:ELBO}). Hence, we need its numerical estimator 
\begin{equation}\label{eq:ELBO_hat}
    \begin{split}
        & \widehat{\ELBO}(\bm X;\theta,\phi) \\
        = & \hat\Ed_{q(\bm Z|\bm X;\phi)}[\ln p(\bm X,\bm Z;\theta) - \ln q(\bm Z|\bm X;\phi)] \\
        = & \frac{1}{K}\sum_{k=1}^K \sbr{\ln p\rbr{\bm X,\bm Z^{(k)};\theta} - \ln q\rbr{\bm Z^{(k)}\middle|\bm X;\phi)}},
    \end{split}
\end{equation}
where $\cbr{\bm Z^{(k)}}_{k=1}^K$ are $K$ Monte Carlo samples from $q(\bm Z|\bm X;\phi)$. The derivative w.r.t. $\theta$ is simple (Appendix.~\ref{appendix:gradient}):
\begin{equation}
    \pderiv{\ELBO(\bm X;\theta,\phi)}{\theta} \approx \pderiv{}{\theta} \widehat{\ELBO}(\bm X;\theta,\phi).
\end{equation}
Since $\bm Z\in\Nd^{T\times H}$ are discrete spike counts from the hidden neurons, the derivative w.r.t. $\phi$ at a particular value $\phi_0$ requires the score function gradient estimator (Appendix.~\ref{appendix:gradient}):
\begin{equation}\label{eq:score}
    \begin{split}
        & \pderiv{\ELBO(\bm X;\theta,\phi)}{\phi} \approx \frac{1}{K}\sum_{k=1}^{K} \bigg\{\left[\ln p\rbr{\bm X,\bm Z^{(k)};\theta}\right. \\
        & \left. - \ln q\rbr{\bm Z^{(k)}\middle|\bm X;\phi_0}\right] \pderiv{}{\phi} \ln q\rbr{\bm Z^{(k)}\middle|\bm X;\phi}\bigg\}.
    \end{split}
\end{equation}
However, previous literature shows that the score function gradient estimator for maximizing ELBO w.r.t. $\phi$ exhibits high variance \citep{paisley2012variational,bengio2013estimating,kingma2013auto,schulman2015gradient}, and hence it could be better to seek a reparameterization trick for sampling $\bm Z^{(k)}$ from $q(\bm Z|\bm X;\phi)$ so that pathwise gradient estimator can be applied. Given there is no reparameterization trick for Poisson distribution, we have to relax the discrete latent variable $\bm Z$ into a continuous variable and reformulate a differentiable POGLM as follows.

\subsection{A Differentiable POGLM}
\paragraph{Relaxation for differentiability.} In this subsection, we formulate a differentiable POGLM via the Gumbel-Softmax distribution \citep{jang2016categorical,maddison2016concrete}. We first set a large enough upper-bound $M$ so that $z_{t,h} \in \cbr{0,1,\dots, M-1}$. Practically, $M$ doesn't need to be very large since the number of spikes in a short enough time bin is limited. Usually, we hope the number of spikes in each time bin is very small (most of them should be 0 or 1) so that the precision of the spike train can be preserved. Without loss of generality, we use $M=5$ in our following experiments. Then, a categorical distribution can be used to approximate the corresponding Poisson distribution:
\begin{equation}\label{eq:categorical}
    z_{t,h} \sim \operatorname{Cat}(\bm \pi(f_{t,h})),
\end{equation}
where
\begin{equation}\label{eq:pi}
    \bm\pi(f_{t,h}) = \scalebox{0.9}{$\displaystyle\rbr{1-\sum_{m=1}^{M-1} \frac{f_{t,h}^m \eu^{f_{t,h}}}{m!},\frac{f_{t,h}^1\eu^{f_{t,h}}}{1!},\dots,\frac{f_{t,h}^{M-1}\eu^{f_{t,h}}}{(M-1)!}}$}
\end{equation}
expands the Poisson distribution truncated $M$. Then, we can use Gumbel-Softmax (GS) to relax the discrete $z_{t,h}$ into a soft one-hot version
\begin{equation}\label{eq:GS}
    \tilde{\bm z}_{t,h} = \rbr{\tilde z_{t,h,0}, \dots, \tilde z_{t,h,M-1}} \sim \operatorname{GS}(\bm\pi(f_{t,h});\tau),
\end{equation}
where $\tilde{\bm z}_{t,h}$ is a soft one-hot vector over a Simplex $\Delta^{M-1} \coloneqq \cbr{\bm z\in[0,1]\middle|\sum_{m=0}^{M-1} z_m = 1}$. Specifically,
\begin{equation}\label{eq:reparameterization}
    \tilde z_{t,h,m} = \frac{\exp[(\ln \pi_m(f_{t,h}) + g_{t,h,m})/\tau]}{\sum_{m'=0}^{M-1} \exp[(\ln \pi_{m'}(f_{t,h}) + g_{t,h,m'})/\tau]},
\end{equation}
where $g_{t,h,m} \overset{\mathrm{i.i.d.}}{\sim} \operatorname{Gumbel}(0,1)$. In practice, we can sample $g$ by sampling $u$ from $\operatorname{Uniform}(0,1)$ and computing $g = -\ln(-\ln(u))$. $\tau > 0$ is a temperature hyperparameter forcing $\tilde{\bm z}_{t,h}$ to be a soft one-hot representation closing to a corner of the Simplex $\Delta^{M-1}$. When $\tau \to 0$, $\tilde{\bm z}_{t,h}$ becomes the hard one-hot representation of the spike count $z_{t,h}$. It is common to choose the temperature $\tau$ in Gumbel-Softmax from [0.1, 1]. If $\tau$ is too large, the relaxation will be too soft; if $\tau$ is too small, numerical issues could arise. In our model, $\tau$ is used to force the soft one-hot close to one corner of the simplex, so we tried $\tau\in\cbr{0.2,0.5,1}$ and found $\tau=0.5$ is an optimal choice that gives good and stable categorical approximation without numerical issue. We fix $\tau=0.5$ in this differentiable model in our experiments, which is a common moderate choice. More details about the Gumbel-Softmax distribution including its likelihood function are in \citet{jang2016categorical,maddison2016concrete}.

\vspace{-0.1in}
\paragraph{Generative and variational model.} 
Given the soft one-hot $\tilde{z}_{t,h,m}$, we next define the equivalent soft hidden spike count as
\vspace{-0.1in}
\begin{equation}\label{eq:equivalent_soft}
    z_{t,h} = \sum_{m=0}^{M-1} m\cdot \tilde z_{t,h,m}.
\end{equation}
Now we are ready to define the complete differentiable generative model $p\rbr{\bm X,\tilde{\bm Z};\theta}$. Visible neurons' spikes $\bm X$ are generated with Eq.~\ref{eq:GLM_v} ($f_{t,v}$) and Eq.~\ref{eq:Poisson_v} (Poisson) where $z_{t,h}$ in Eq.~\ref{eq:GLM_v} is now defined by the above Eq.~\ref{eq:equivalent_soft}. Hidden variables $\tilde{\bm Z}$ are generated from Eq.~\ref{eq:GLM_h} ($f_{t,h}$) and Eq.~\ref{eq:GS} (GS), instead of Eq.~\ref{eq:GLM_h} and Eq.~\ref{eq:Poisson_h} (Poisson). Similarly the sampling process of $\tilde{\bm Z}$ in the variational model $q\rbr{\tilde{\bm Z}|\bm X;\phi}$ changes from Eq.~\ref{eq:homo_Poisson} ($f_{t,h}$) and Eq.~\ref{eq:Poisson_h} (Poisson) to Eq.~\ref{eq:homo_Poisson} and Eq.~\ref{eq:GS} (GS). Now, the complete spike train of this differentiable POGLM is $\cbr{\bm X,\tilde{\bm Z}}$ where $\tilde{\bm Z}\in \rbr{\Delta^{M-1}}^{T\times H} \subsetneqq [0,1]^{T\times H\times M}$.

\vspace{-0.1in}
\paragraph{Pathwise gradient estimator.} With both the generative model and the variational model differentiable, the pathwise gradient estimator can be used for optimizing $\phi$. Particularly, by the reparameterization trick in Eq.~\ref{eq:reparameterization} abbreviated as $\tilde{\bm Z}|\bm X;\phi = r(\bm G|\bm X;\phi)$ where $\bm G\sim \operatorname{Gumbel}(\bm G;0,1)$, we have the transformation relationship $q\rbr{\tilde{\bm Z}\middle|\bm X;\phi}\ \du \tilde{\bm Z} = \operatorname{Gumbel}(\bm G;0,1)\ \du \bm G$ \citep{schulman2015gradient}. Then, the pathwise gradient estimator of the derivative of ELBO w.r.t. $\phi$ is:
\vspace{-0.1in}
\begin{equation}\label{eq:pathwise}
    \pderiv{\ELBO(\bm X;\theta,\phi)}{\phi} \approx \pderiv{}{\phi}\widehat{\ELBO}(\bm X;\theta,\phi),
\vspace{-0.1in}
\end{equation}
and
\vspace{-0.1in}
\begin{equation}
    \begin{split}
        & \widehat{\ELBO}(\bm X;\theta,\phi) \\
        = & \frac{1}{K}\sum_{k=1}^K \sbr{\ln p\rbr{\bm X,\tilde{\bm Z}^{(k)};\theta)} - \ln q\rbr{\tilde{\bm Z}^{(k)}\middle|\bm X;\phi}},
    \end{split}
\vspace{-0.1in}
\end{equation}
where $\tilde{\bm Z}^{(k)} = r(\bm G^{(k)}|\bm X;\phi)$ and $\cbr{\bm G^{(k)}}_{k=1}^K$ are $K$ Monte Carlo samples from $\operatorname{Gumbel}(\bm G;0,1)$. The detailed derivation of this is shown in Appendix.~\ref{appendix:gradient}.

\vspace{-0.05in}
\paragraph{Relax to general continuous distributions.} In fact, the differentiable POGLM introduced above is already compatible with any continuous distributions that satisfy the following two requirements: (1) the distribution is parameterized by a single mean parameter, since the GLM structure provides a single firing rate $f_{t,h}$ representing the mean statistic of the (equivalent soft) spike count $z_{t,h}$; and (2) a reparameterization trick should exist for sampling such a distribution. For example, in the generative and variational models, we can assume a soft hidden spike count is from an exponential distribution
\vspace{-0.1in}
\begin{equation}\label{eq:exp}
    z_{t,h} \sim \operatorname{Exp}(1/f_{t,h})
\end{equation}
with mean $f_{t,h}$ computed from Eq.~\ref{eq:GLM_h} and \ref{eq:homo_Poisson}, by sampling
\vspace{-0.1in}
\begin{equation}
    z_{t,h} = -f_{t,h} \ln(1-u),\quad u\sim \operatorname{Unif}(0, 1).
\vspace{-0.1in}
\end{equation}
Compared with the GS distribution in which the equivalent soft hidden spike count is close to an integer in $\cbr{0,1,\dots,M-1}$, $z_{t,h}$ from the exponential distribution can be any value in $\Rd_{\geqslant 0}$. More details about the possible choices of the distributions are in Sec.~\ref{sec:3} and Fig.~\ref{fig:different_distributions}.

\vspace{-0.05in}
\subsection{Sampling scheme of the variational model}\label{sec:sampling_schemes}
So far, we have proposed the differentiable POGLM to resolve the first issue mentioned in Sec.~\ref{sec:1}. Now, we turn to the second issue---the choice of the variational model. Specifically, we need to design the formula for $f_{t,h}$ in the variational model. Clearly, the homogeneous one we introduced in Eq.~\ref{eq:homo_Poisson} is oversimplified so that the variation distribution family of $q(\bm Z|\bm X;\phi)$ is very far from the posterior distribution $p(\bm Z|\bm X;\theta)$. A good choice of the variational distribution family that is much closer to the true posterior distribution is critical to the success of VI. Here, we discuss five candidates as follows:\\
$\bullet$ \textbf{Homogeneous Poisson}: $f_{t,h} = \sigma(c_h),\ \forall t\in \cbr{1,\dots,T}$, and the variational parameter set is $\phi = \cbr{\bm c_H}$, where $\bm c_H = [c_1,\dots, c_H]^\Tu$. However, this is too simple to serve as a variational distribution family in practice.\\
$\bullet$ \textbf{Inhomogeneous Poisson (mean-field)}: $f_{t,h} = \sigma(c_{t,h})$, and $\phi = \cbr{\bm C_{T\times H}\in \Rd^{T\times H}}$. Although the mean-field theory is widely applicable to a lot of latent variable models, it lacks the dependency on the visible spike train $\bm X$. For POGLM, this learned $\phi$ is bonded to the training spike train $p(\bm Z_{\mathrm{train}}|\bm X_{\mathrm{train}};\theta) \approx q(\bm Z_{\mathrm{train}}|\bm X_{\mathrm{train}};\phi)$, but unable to be generalized to the test spike train $p(\bm Z_{\mathrm{test}}|\bm X_{\mathrm{test}};\theta)$. Besides, both homogeneous and inhomogeneous Poisson have no message passing between neurons, and hence are very unhelpful for learning the neural connection matrix $\bm W$.\\
$\bullet$ \textbf{Forward-self} \citep{jimenez2014stochastic,kajino2021differentiable}: A typical and intuitive way is to assume the true posterior distribution $p(\bm Z|\bm X;\theta)$ can be approximated by a variational distribution $q(\bm Z|\bm X;\phi)$ which is also a GLM w.r.t. $\bm Z$ where $\bm X$ is fixed (Fig.~\ref{fig:graphical_model}(b)), i.e.,
\vspace{-0.1in}
\begin{equation}\label{eq:forward_self}
    \begin{split}
        f_{t,h} = & \sigma\Bigg(c_h + \sum_{v'=1}^{V} a_{h\gets v'}\cdot \rbr{\sum_{l=1}^L x_{t-l,v'}\ \psi_l} \\
        & + \sum_{h'=1}^H a_{h\gets h'} \cdot \rbr{\sum_{l=1}^L z_{t-l,h'}\ \psi_l} \Bigg),
    \end{split}
\vspace{-0.1in}
\end{equation}
and $\phi = \cbr{\bm c_H, \bm A}$. Particularly,
\vspace{-0.1in}
\begin{equation}
    \bm A = \begin{bmatrix}
        \bm O_{V\gets V} & \bm O_{V\gets H} \\
        \bm A_{H\gets V} & \bm A_{H\gets H}
    \end{bmatrix} \in \Rd^{N\times N}.
\vspace{-0.1in}
\end{equation}
The top two blocks $\bm O_{V\gets V}, \bm O_{V\gets H}$ are all zeros since we don't need to sample the visible spike train $\bm X$. $\bm A_{H\gets V}$ and $\bm A_{H\gets H}$ represent the visible-to-hidden and hidden-to-hidden influences. To use Eq.~\ref{eq:forward_self}, we need to sample from $t=1$ to $t=T$ sequentially, since the current sample $z_{t,h}$ relies on the previous samples $\bm Z_{t-L:t-1,1:H}$.\\
$\bullet$ \textbf{Forward}: Due to the low efficiency of the forward-self sampling process, an easier alternative approach is to eliminate the hidden-to-hidden block (i.e., the third term in Eq.~\ref{eq:forward_self}). Then we get the forward message passing scheme (Fig.~\ref{fig:graphical_model}(c)):
\vspace{-0.1in}
\begin{equation}\label{eq:forward}
    f_{t,h} = \sigma\rbr{c_h + \sum_{v'=1}^{V} a_{h\gets v'}\cdot \rbr{\sum_{l=1}^L x_{t-l,v'}\ \psi_l}},
\vspace{-0.1in}
\end{equation}
and $\phi = \cbr{\bm c_H,\bm A}$. Now,
\vspace{-0.1in}
\begin{equation}
    \bm A = \begin{bmatrix}
        \bm O_{V\gets V} & \bm O_{V\gets H} \\
        \bm A_{H\gets V} & \bm O_{H\gets H}
    \end{bmatrix} \in \Rd^{N\times N}.
\vspace{-0.1in}
\end{equation}
The forward variational distribution can be sampled in a parallel style, since $z_{t,h}$ are no longer conditioned on each other. Note that eliminating the hidden-to-hidden block could omit the factor of hidden-to-hidden influences theoretically. But in fact, it is very challenging to learn the actual hidden-to-hidden influences $\bm W_{H\gets H}$ practically given the long sequential sampling procedure, especially when $V > H$ under most realistic model assumptions.\\
$\bullet\enspace$\textbf{Forward-backward}: Both of the previous two sampling schemes omit mimicking an important relationship---the hidden-to-visible influences $\bm W_{V\gets H}$ in the generating process $p(\bm X,\bm Z;\theta)$. Therefore, we introduce the forward-backward message passing scheme (Fig. \ref{fig:graphical_model}(d)),
\vspace{-0.1in}
\begin{equation}\label{eq:forward_backward}
    \begin{split}
        f_{t,h} = & \sigma\Bigg(c_n + \sum_{v'=1}^{V} a_{h\gets v'}\cdot \rbr{\sum_{l=1}^L x_{t-l,v'}\ \psi_l} \\
        & + \sum_{v'=1}^V a_{v'\gets h} \cdot \rbr{\sum_{l=1}^L x_{t+l,v'}\ \psi_l} \Bigg),
    \end{split}
\vspace{-0.1in}
\end{equation}
and $\phi = \cbr{\bm c_H,\bm A}$. Now,
\vspace{-0.1in}
\begin{equation}
    \bm A = \begin{bmatrix}
        \bm O_{V\gets V} & \bm A_{V\gets H} \\
        \bm A_{H\gets V} & \bm O_{H\gets H}
    \end{bmatrix} \in \Rd^{N\times N},
\vspace{-0.1in}
\end{equation}
where the $\bm A_{V\gets H}$ block mimics the hidden-to-visible influences $\bm W_{V\gets H}$ in the generating process $p(\bm X,\bm Z;\theta)$, via including the contribution from future visible spikes $X_{t+1:t+L,1:V}$ into sampling the current $z_{t,h}$ (i.e., the third term in Eq.~\ref{eq:forward_backward}).

\begin{figure}
\vspace{-0.05in}
    \centering
    \includegraphics[width=\linewidth]{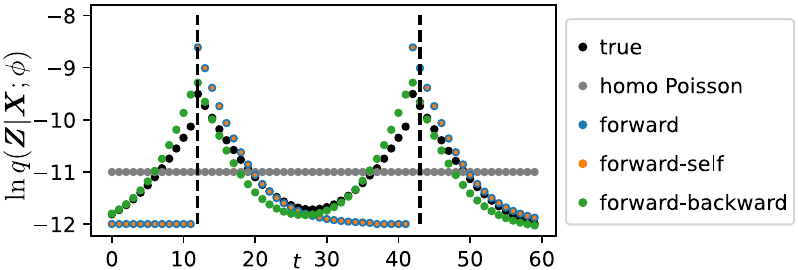}
    \vspace{-0.2in}
    \caption{An example of comparison among different variational distributions $q(\bm Z|\bm X;\phi)$ with the true posterior $p(\bm Z|\bm X;\theta)$ (black dots). There is one visible neuron and one hidden neuron. Two visible spikes from the visible neuron happen at the dashed lines. Different dotted curves represent the approximated log-likelihood of one hidden spike happening at different time bins. Only the forward-backward recapitulates the true distribution. The forward and forward-self miss the uprising trends before the two observed spikes, due to lack of a back-propagated message.}
    \label{fig:variational_distributions}
    \vspace{-0.1in}
\end{figure}

Fig.~\ref{fig:variational_distributions} visually compares different variational distributions, helping us understand the superiority of the forward-backward sampling, which excels in approximating the true posterior distribution.

\vspace{-0.1in}
\section{Experiments}\label{sec:3}
For a comprehensive analysis and comparison, we consider the \textbf{method combinations} of different \textbf{inference methods} $\times$ different \textbf{variational sampling schemes}.

\begin{figure}[t]
\vspace{-0.05in}
    \centering
    \includegraphics[width=\linewidth]{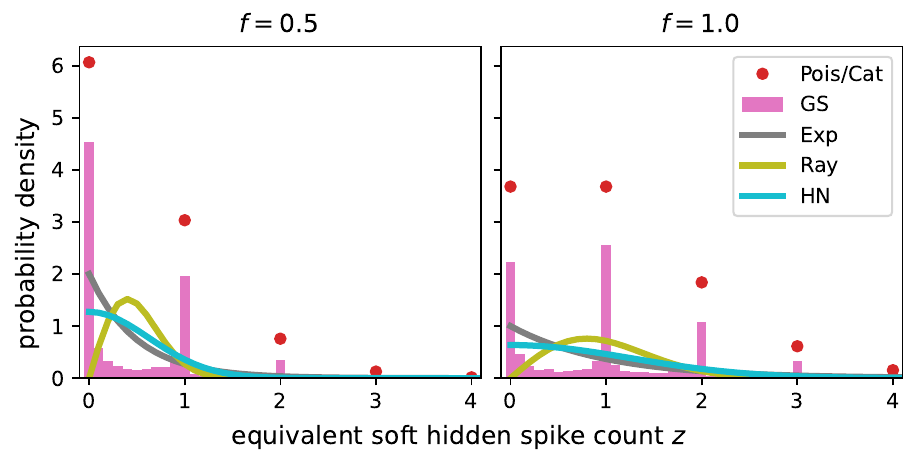}
    \vspace{-0.3in}
    \caption{A visualization of different choices of the soft hidden spike count distribution, under firing rate $f=0.5$ and $f=1.0$. Most of $z$ from GS approximating the original Poisson distribution are close to integer points, but the three continuous distributions (Exp, Ray, and HN) are not.}
    \label{fig:different_distributions}
    \vspace{-0.2in}
\end{figure}

\vspace{-0.05in}
\paragraph{Inference methods.} We consider seven inference methods. Each of them is identified by a distribution of the hidden spike and the gradient estimator used in VI.\\
$\bullet$ \textbf{Poisson (Pois)}: This is the original POGLM (Eq.~\ref{eq:GLM_v}, \ref{eq:Poisson_v}, \ref{eq:GLM_h}, \ref{eq:Poisson_h}, \ref{eq:homo_Poisson}). This is named after the original Poisson distribution of the hidden spike train $\bm Z$ of the original POGLM. Since this is a discrete distribution, only the score function gradient estimator can be adopted in VI.\\
$\bullet$ \textbf{Categorical (Cat)}: This is the first intermediate model between the original POGLM and the differentiable POGLM, where we don't use Gumbel-Softmax in Eq.~\ref{eq:GS} to approximate but keep the categorical distribution (Eq.~\ref{eq:categorical}). Similar to Poisson, only the score function gradient estimator can be adopted in VI.\\
$\bullet$ \textbf{Gumbel-Softmax-score (GS-s)}: This is the differentiable POGLM with GS (Eq.~\ref{eq:GS}) as the soft hidden spike count distributions, but we still use the score function gradient estimator (Eq.~\ref{eq:score}) when updating $\phi$ although this model is already differentiable.\\
$\bullet$ \textbf{Gumbel-Softmax-pathwise (GS-p)}: This is the differentiable POGLM with GS (Eq.~\ref{eq:GS}) as the soft hidden spike count distributions, where we use the pathwise gradient estimator (Eq.~\ref{eq:pathwise}) when updating $\phi$. This is the inference method we expect to perform better than the previous three. To experiment with the generalization from GS to other single-parameter continuous distributions, we try the following three and use the pathwise gradient estimator.\\
$\bullet$ \textbf{Exponential (Exp)}: Eq.~\ref{eq:exp}. The probability density function (likelihood) is $\Pd[z;f] = \frac{1}{f}\exp\rbr{-fz}$.\\
$\bullet$ \textbf{Rayleigh (Ray)}: $z_{t,h}\sim \operatorname{Ray}\rbr{\sqrt{\frac{2}{\uppi}}f_{t,h}}$. The probability density function is $\Pd[z;f] = \frac{\uppi z}{2f^2}\exp\rbr{-\frac{\uppi z^2}{4f^2}}$.\\
$\bullet$ \textbf{Half-normal (HN)}: $z_{t,h}\sim \operatorname{HN}\rbr{\sqrt{\frac{\uppi}{2}}f_{t,h}}$ The probability density function is $\Pd[z;f] = \frac{2}{\uppi f}\exp\rbr{-\frac{z^2}{\uppi f^2}}$.\\
The aim of including the two intermediate models (Cat and GS-s) is to change the model from the original POGLM with Poisson as the hidden spike count distribution and the score function gradient estimator to the differentiable POGLM step-by-step with GS as the hidden spike count distribution and the pathwise gradient estimator. Through this controlled variable design, we can have a better understanding of the final differentiable POGLM with continuous soft hidden spike count distributions. Since we have no prior knowledge about which single parameter distribution is better, we try the three common ones: Exp, Ray, and HN. A visualization of these distributions is shown in Fig.~\ref{fig:different_distributions}.

\begin{figure*}[!ht]
    \centering
    \includegraphics[width=\textwidth]{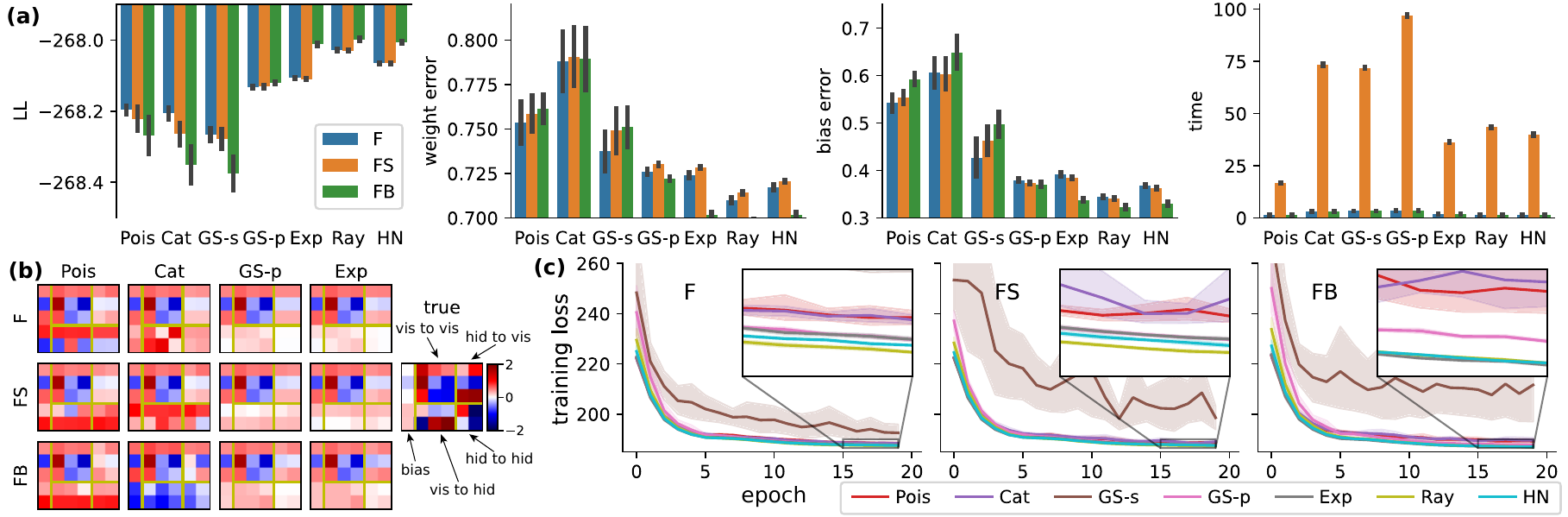}
    \vspace{-0.25in}
    \caption{\textbf{(a)}: The test log-likelihood (LL) on the test set, the weight error, the bias error, and the running time of different method combinations. \textbf{(b)}: An example of the learned weight matrix and bias vector compared with the true of selected method combinations. Visualization of all method combinations is in Fig.~\ref{fig:synthetic_weight} in Appendix.~\ref{appendix:supplementary_figures} \textbf{(c)}: The learning curves of different method combinations.}
    \label{fig:synthetic}
    \vspace{-0.15in}
\end{figure*}

\vspace{-0.1in}
\paragraph{Variational sampling schemes.} We consider three sampling schemes.\\
$\bullet$ \textbf{Forward (F)}: The sampling scheme illustrated in Eq.~\ref{eq:forward_self} and Fig.~\ref{fig:graphical_model}(c).\\
$\bullet$ \textbf{Forward-self (FS)}: The sampling scheme illustrated in Eq.~\ref{eq:forward} and Fig.~\ref{fig:graphical_model}(b).\\
$\bullet$ \textbf{Forward-backward (FB)}: The sampling scheme illustrated in Eq.~\ref{eq:forward_backward} and Fig.~\ref{fig:graphical_model}(d).\\
The homogeneous and inhomogeneous Poisson will be ignored in the following experiments due to their over-simplicity or incompatibility.

The original approach to solving POGLM can be viewed as the combination of Poisson $\times$ forward-self \citep{pillow2007neural,jimenez2014stochastic,linderman2017bayesian}. Our newly proposed differentiable POGLM with GS and other continuous distributions combined with the FB message-passing scheme should be the optimal combinations we expect. For clarity, Appendix.~\ref{appendix:method_combinations} provides a comprehensive summary of these method combinations.

\vspace{-0.1in}
\paragraph{Evaluation.} Although we have different inference methods when evaluating log-likelihood (LL) on the test dataset, all inference methods are set back to the original POGLM form, where Poisson log-likelihood (Eq.~\ref{eq:Poisson_v} and Eq.~\ref{eq:Poisson_h}) is adopted for a fair comparison. The LL metric can be used on both synthetic and real-world datasets. Furthermore, since we are also very concerned about the parameter recovery, as we illustrated in Sec.~\ref{sec:1} that naively apply VI to solve the POGLM directly cannot obtain an ideal parameter recovery result due to the difficulty of the POGLM problem itself, we will compare the average absolute error of the estimated parameter on the synthetic dataset.

\vspace{-0.05in}
\subsection{Synthetic Dataset}
\vspace{-0.1in}
The synthetic dataset aims to compare different method combinations comprehensively. With the known true parameters, we can validate that better performance corresponds to smaller parameter errors. We can also check the benefit of applying the pathwise gradient estimator.

\vspace{-0.1in}
\paragraph{Dataset.} We randomly create 10 sets of parameters for generating the synthetic datasets, wherein each set $w_{n\gets n'}\overset{\mathrm{i.i.d.}}{\sim}\operatorname{Unif}(-2, 2)$ and $b_n\overset{\mathrm{i.i.d}}{\sim}\operatorname{Unif}(-0.5,0.5)$. Each set corresponds to a trial, resulting in a total of 10 trials. The model consists of 5 neurons, with the first 3 being visible and the remaining 2 being hidden. For each trial, we generate 40 spike trains for training and 20 spike trains for testing. Each spike train has 100 time bins.

\vspace{-0.1in}
\paragraph{Experimental setup.} The initial values for the linear weights and biases of the model used for learning are also randomly initialized as above. We utilize the Adam optimizer \citep{kingma2014adam} with a learning rate of 0.05. The optimization process runs for 20 epochs, and within each epoch, optimization is performed using 4 batches, each of size 10. The entire process is repeated 10 times with different random seeds for each trial, and the performance and the error bar are reported based on these repetitions.

\begin{figure*}[!ht]
    \centering
    \includegraphics[width=\textwidth]{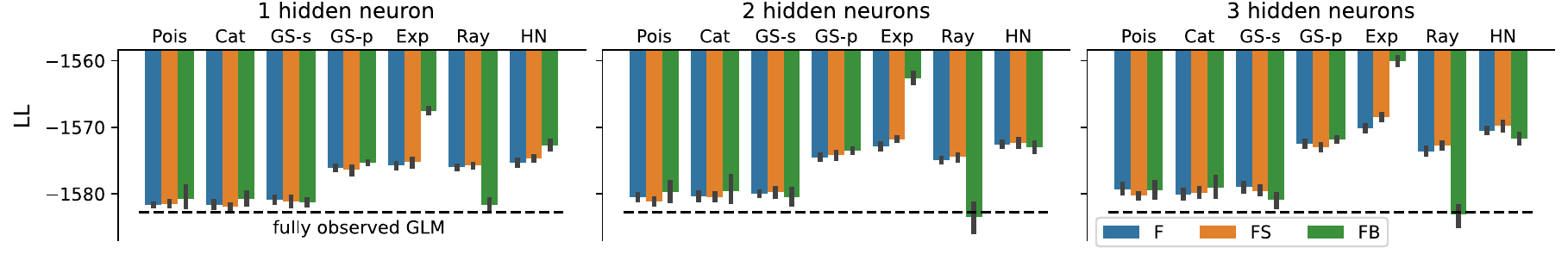}
    \vspace{-0.25in}
    \caption{The test log-likelihood (LL) of different method combinations under $H\in\cbr{1,2,3}$ hidden neurons. The dashed black line represents the test LL of the fully observed GLM as the baseline.}
    \label{fig:rgc_bar}
    \vspace{-0.15in}
\end{figure*}

\vspace{-0.05in}
\paragraph{Results.}
In Fig.~\ref{fig:synthetic}(a), we can see that for each sampling scheme, the test LL drops when we use categorical and then GS to approximate the discrete Poisson distribution. However, when we change the gradient estimator from the score function to the pathwise, the LL increases significantly for GS and exceeds the original Poisson. The LL keeps increasing when we change the distribution from GS to Exp, Ray, and HN. This implies that although the hidden spikes are generated from the Poisson distribution, the posterior distribution might not be Poisson, but a discrete distribution that is closer to a continuous distribution like the exponential in shape. From the view of different sampling schemes, FB becomes better than F and FS when a good inference method is used, e.g., GS-p or Exp. Without a good inference method, the forward sampling scheme performs better due to its simplicity. In summary, a differentiable POGLM using pathwise gradient estimator $\times$ the FB sampling scheme promises a better performance.

The weight error and bias error in Fig.~\ref{fig:synthetic}(a) quantitatively validate that a better LL corresponds to a smaller parameter error. Consistent with the LL bar plot, the weight error and bias error bar plots indicate that Exp, Ray, and HN are better than GS-p and better than the remaining. With a continuous distribution and the pathwise gradient estimator as a differentiable method, FB is the optimal sampling scheme. Fig.~\ref{fig:synthetic}(b) visualizes the recovered weight matrix and bias vector of some selected method combinations versus the true one used for generating the dataset. The differences between these recovered weights and biases are mainly from the hidden related weight blocks and hidden biases.

In addition to the performance, we also compare the running time of different method combinations in Fig.~\ref{fig:synthetic}(a). Due to the sequential dependency of the FS, the running time is significantly longer than F and FB. This implies the benefit of excluding the complicated ``self'' message-passing component in the variational model. Besides, it takes a bit longer to convert a discrete Poisson distribution to its hard (Cat) and soft (GS) approximating distribution.

Fig.~\ref{fig:synthetic}(c) plots the loss curves of different inference methods using different sampling schemes. The loss curves of those differentiable inference methods (GS-p, Exp, Ray, and HN) are smoother than the others. The small error bars of these differentiable inference methods imply the smaller variance of the pathwise gradient estimator when updating the variational model parameter $\phi$ than that of the score function gradient estimator.

\vspace{-0.05in}
\subsection{Retinal Ganglion Cell (RGC) Dataset}
This dataset aims to understand the performances of different method combinations on a real-world dataset, and the interpretability of the estimated model parameters.

\vspace{-0.1in}
\paragraph{Dataset.} Next, we apply various method combinations to analyze a real neural spike train recorded from 27 basal ganglion neurons while a mouse is engaged in a visual task for approximately 20 minutes \citep{pillow2012fully}. Specifically, neurons 1-16 are OFF cells, while neurons 17-27 are ON cells.

\vspace{-0.1in}
\paragraph{Experimental setup.} We partition the spike train into training and test sets, using the first 2/3 segment for training and the remaining 1/3 segment for testing. The original spike train is binned into spike count via 50 ms time bins. To facilitate the application of the stochastic gradient descent algorithm, we divide the entire sequence into 14400 pieces in total, each consisting of 100 time bins. Since we do not have prior knowledge of how many hidden neurons should be assumed, we initially train a fully observed GLM as a baseline. Subsequently, we assume the presence of $H\in\cbr{1,2,3}$ hidden representative neurons and train the model using different method combinations. The optimization is performed using the Adam optimizer with a learning rate of 0.02. Each training procedure undergoes 20 epochs, employing a batch size of 32. We repeat the training and test process 10 times with different random seeds and report the performance and error bar.

\vspace{-0.1in}
\paragraph{Results.}Similar to the synthetic dataset, we plot the test LL of different method combinations under 1, 2, and 3 hidden neurons in Fig.~\ref{fig:rgc_bar}. Besides, a fully observed GLM is also trained as a baseline for comparison. First of all, with the assumption of containing hidden neurons, the performances of different method combinations all become better, except for the Ray $\times$ FB. Second, differentiable inference methods are better than non-differentiable inference methods in general. Particularly, Exp $\times$ FB is significantly better than all other method combinations. Third, for most of the method combinations, increasing the number of hidden neurons improves the LL, especially for GS-p and Exp.

\begin{figure}[!ht]
\vspace{-0.1in}
    \centering
    \includegraphics[width=\linewidth]{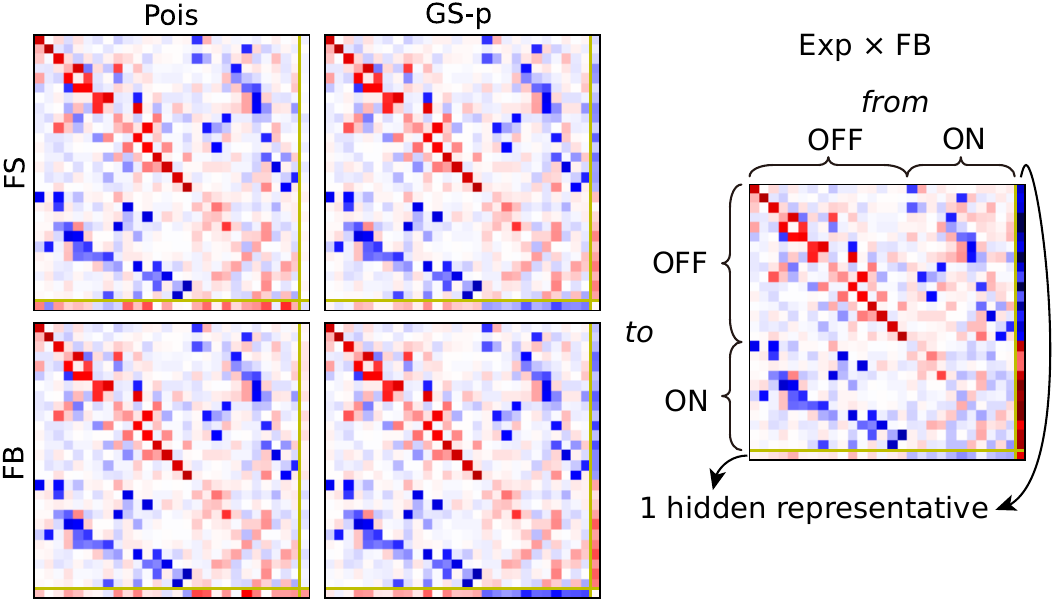}
    \vspace{-0.25in}
    \caption{The learned weight matrix of selected method combinations. Visualization of all method combinations is in Fig.~\ref{fig:rgc_weight} in Appendix.~\ref{appendix:supplementary_figures}.}
    \label{fig:rgc_selected_weight}
    \vspace{-0.1in}
\end{figure}

\begin{figure*}[t]
\vspace{-0.05in}
    \centering
    \includegraphics[width=\textwidth]{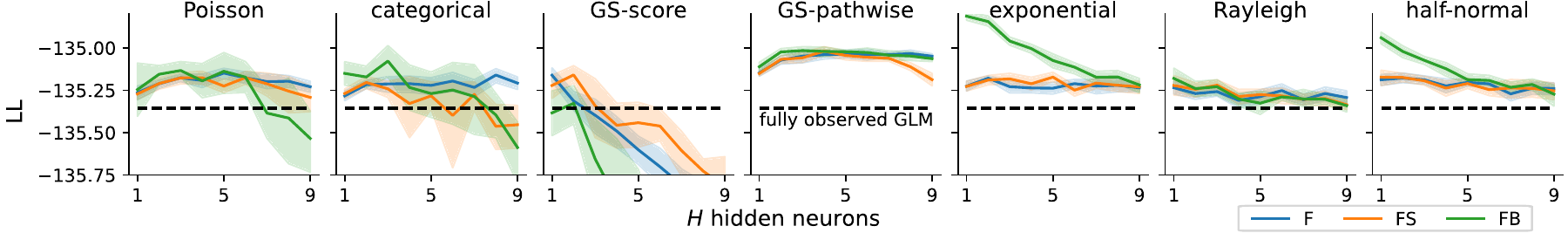}
    \vspace{-0.3in}
    \caption{The curves of the test log-likelihood (LL) v.s. the number of hidden neurons $H$, for different method combinations.}
    \label{fig:pvc-5}
    \vspace{-0.15in}
\end{figure*}

In addition to knowing that the differentiable inference methods $\times$ FB are better than others, we are also curious about the interpretation of the learned model parameters. Take 1 hidden neuron as an example, Fig.~\ref{fig:rgc_selected_weight} shows that, the learned one hidden representative by Exp $\times$ FB serves as a negative feedback regulating unit. Particularly, the weights from this hidden representative to all OFF cells are negative, and to all ON cells are positive; the weights from nearly all OFF cells to this hidden representative are positive, and from all OFF cells are negative. The signs of the weights from this hidden representative to visible neurons provide clear indications of the type of visible neurons. Similar but weaker results can also be obtained by GS-p but not by Poisson.

\vspace{-0.05in}
\subsection{The PVC-5 Dataset}
This dataset aims to investigate the performance variation of different method combinations w.r.t. different numbers of hidden neurons.
\vspace{-0.1in}
\paragraph{Dataset.} Finally, we apply different method combinations to a dataset obtained from the primary visual cortex (PVC-5) \cite{chu2014multi}\footnote{\url{https://crcns.org/data-sets/pvc/pvc-5}}. This dataset consists of recordings from the primary visual cortex (V1) of a macaque monkey over a 15-minute duration without the presentation of any stimuli. Three adjacent neurons were recorded through contact with an electrode.

\vspace{-0.1in}
\paragraph{Experimental setup.} Similarly to the RGC dataset, we train the model on the initial 7.5 minutes of data and evaluate the test log-likelihood on the subsequent 7.5 minutes. The original spike train is binned into spike count via 20 ms time bins. The training set is divided into 225 pieces equally and the batch size for training is 25. Since there are only three visible neurons, we can try more numbers of hidden neurons $H\in\cbr{1,\dots,9}$ and understand the change of performance w.r.t. the number of hidden neurons, especially when $H\gg V$. The optimization is performed using the Adam optimizer for 20 epochs with a learning rate of 0.1. We repeat the training and test process 10 times with different random seeds and report the performance and error bar.

\vspace{-0.1in}
\paragraph{Results.} Fig.~\ref{fig:pvc-5} shows the performance of different method combinations w.r.t. number of hidden neurons. No matter what method combination we choose, the optimal number of hidden neurons is no more than 3. This means assuming more hidden neurons might not guarantee an improvement of the performance but is likely to introduce redundancy and result in a dropped performance. With more hidden neurons, the performance of those non-differentiable inference methods becomes even worse than a fully observed GLM.

Among all method combinations, Exp $\times$ FB with less than 3 hidden neurons seems to be the best. GS-p (with all three sampling schemes) is more robust to different numbers of hidden neurons than other inference methods. Furthermore, the variance of the test LL of those differentiable inference methods is much smaller than that of those non-differentiable inference methods, due to the benefit from using the pathwise gradient estimator.

\vspace{-0.05in}
\section{Related Works}
\vspace{-0.05in}
Some previous works consider the POGLM in its point process form, i.e., a generalized multivariate partially observable Hawkes process, in which spike trains are not binned into spike count but keep the form of spiking event timestamps. For example, \citet{zhou2021multivariate,shelton2018hawkes,mei2019imputing} treated the problem as missing data (missing all the spiking data from hidden neurons); \citet{kajino2021differentiable} proposed a differentiable point process model to enable the use of the pathwise gradient estimator.

Through the point process, however, the data structure that stores the spike timestamps is usually not ideal \citep{xu2018poppy}. The detailed reasons are illustrated in Appendix.~\ref{appendix:point_process}. In this paper, we only focus on POGLM, i.e., pre-binned spike count trains, and more detailed discussions regarding the relationship between the (discrete) GLM and the (continuous) generalized Hawkes process can be found in Appendix.~\ref{appendix:point_process}.

\vspace{-0.05in}
\section{Conclusion}
\vspace{-0.05in}
In this paper, we propose a differentiable version of the partially observable generalized linear model (POGLM), in which the pathwise gradient estimator becomes applicable when doing variational inference (VI). Due to the inexpressivity and low sampling efficiency of the existing forward-self sampling scheme, we propose the new forward-backward message-passing sampling scheme, enhance the message passing from hidden neurons to visible neurons, and result in a better variational model for VI. Comprehensive comparisons between different method combinations on one synthetic and two real-world shows that a differentiable inference method with the forward-backward sampling scheme could produce a higher likelihood on the test set and better parameter recovery.

Note that the relaxation from the Gumbel-Softmax distribution to general continuous distributions loses the meaning of $Z$ as representing spike count, but can produce better performance. It is interesting but challenging to investigate whether a general continuous distribution is closer to the true posterior distribution than the discrete Poisson distribution. This limitation is a big topic that could be a future direction.







\section*{Impact Statements}
This paper presents work whose goal is to advance the field of Machine Learning. There are many potential societal consequences of our work, none of which we feel must be specifically highlighted here.


\bibliography{ref}
\bibliographystyle{icml2024}

\newpage
\appendix
\onecolumn
\section{Appendix}
\subsection{Gradient estimators of ELBO}\label{appendix:gradient}
Here we provide detailed derivations for the gradient estimators w.r.t. $\theta$ and $\phi$ of $\ELBO(\bm X;\theta,\phi)$. The derivative w.r.t. $\theta$ is:
\begin{equation}
    \begin{split}
        \pderiv{\ELBO(\bm X;\theta,\phi)}{\theta} = & \frac{1}{\abs{\Nd^{T\times H}}}\sum_{\bm Z\in \Nd^{T\times H}} q(\bm Z|\bm X;\phi) \pderiv{}{\theta}\sbr{\ln p(\bm X,\bm Z;\theta) - \ln q(\bm Z|\bm X;\phi)} \\
        \approx & \frac{1}{K}\sum_{k=1}^K \pderiv{}{\theta} \sbr{\ln p\rbr{\bm X,\bm Z^{(k)};\theta} - \ln q\rbr{\bm Z^{(k)}\middle|\bm X;\phi}} \\
        = & \pderiv{}{\theta} \widehat{\ELBO}(\bm X;\theta,\phi).
    \end{split}
\end{equation}
The score function gradient estimator (i.e., Eq.~\ref{eq:score}) of the derivative of ELBO w.r.t. $\phi$ at a particular value $\phi_0$ is:
\begin{equation}
    \begin{split}
        \pderiv{\ELBO(\bm X;\theta,\phi)}{\phi} = & \frac{1}{\abs{\Nd^{T\times H}}} \sum_{\bm Z\in \Nd^{T\times H}} \pderiv{}{\phi}q(\bm Z|\bm X;\phi) \sbr{\ln p(\bm X,\bm Z;\theta) - \ln q(\bm Z|\bm X;\phi_0)} \\
        & + q(\bm Z|\bm X;\phi_0) \pderiv{}{\phi}\sbr{\ln p(\bm X,\bm Z;\theta) - \ln q(\bm Z|\bm X;\phi)} \\
        = & \frac{1}{\abs{\Nd^{T\times H}}} \sum_{\bm Z\in \Nd^{T\times H}}\sbr{\ln p(\bm X,\bm Z;\theta) - \ln q(\bm Z|\bm X;\phi_0)} q(\bm Z|\bm X;\phi) \pderiv{}{\phi} \ln q(\bm Z|\bm X;\phi) \\
        & - \frac{1}{\abs{\Nd^{T\times H}}} \sum_{\bm Z\in \Nd^{T\times H}} \pderiv{}{\phi} q(\bm Z|\bm X;\phi) \\
        \approx & \frac{1}{K}\sum_{k=1}^{K} \sbr{\ln p\rbr{\bm X,\bm Z^{(k)};\theta} - \ln q\rbr{\bm Z^{(k)}\middle|\bm X;\phi_0}} \pderiv{}{\phi} \ln q\rbr{\bm Z^{(k)}\middle|\bm X;\phi} - 0.
    \end{split}
\end{equation}
The pathwise gradient estimator (i.e., Eq.~\ref{eq:pathwise}) of the derivative of ELBO w.r.t. $\phi$ is:
\begin{equation}
    \begin{split}
        \pderiv{\ELBO(\bm X;\theta,\phi)}{\phi} = & \pderiv{}{\phi} \int_{\tilde{\bm Z}} q\rbr{\tilde{\bm Z}\middle|\bm X;\phi} \sbr{\ln p\rbr{\bm X,\tilde{\bm Z};\theta} - \ln q\rbr{\tilde{\bm Z}\middle|\bm X;\phi_0}}\ \du \tilde{\bm Z} \\
        = & \pderiv{}{\phi} \int_{\tilde{\bm Z}} \operatorname{Gumbel}(\bm G;0,1) \sbr{\ln p\rbr{\bm X,r(\bm G|\bm X;\phi);\theta} - \ln q(r(\bm G|\bm X;\phi)|\bm X;\phi)} \ \du \bm G \\
        \approx & \pderiv{}{\phi} \sum_{k=1}^K \sbr{\ln p\rbr{\bm X,r\rbr{\bm G^{(k)}\middle|\bm X;\phi};\theta} - \ln q\rbr{r\rbr{\bm G^{(k)}\middle|\bm X;\phi}\middle|\bm X;\phi}} \\
        = & \pderiv{}{\phi} \widehat{\ELBO}(\bm X;\theta,\phi).
    \end{split}
\end{equation}

\clearpage
\subsection{Method combinations}\label{appendix:method_combinations}
We summarize different method combinations in this subsection.

\subsubsection{Generative procedure}
For the firing rate $f_{t,v}$ of a visible neuron $v$ and the firing rate $f_{t,h}$ of a hidden neuron $h$ at time $t$:
\begin{equation}
    \begin{cases}
        f_{t,v} = \sigma\rbr{b_v + \sum_{v'=1}^V w_{v\gets v'} \cdot \rbr{\sum_{l=1}^L x_{t-l,v'}\ \psi_l} + \sum_{h'=1}^H w_{v\gets h'} \cdot \rbr{\sum_{l=1}^L z_{t-1,h'}\ \psi_l}}, \\
        f_{t,h} = \sigma\rbr{b_h + \sum_{v'=1}^V w_{h\gets v'} \cdot \rbr{\sum_{l=1}^L x_{t-l,v'}\ \psi_l} + \sum_{h'=1}^H w_{h\gets h'} \cdot \rbr{\sum_{l=1}^L z_{t-1,h'}\ \psi_l}}.
    \end{cases}
\end{equation}
The parameter set is $\theta = \cbr{\bm b,\bm W}$, where $\bm b = \begin{bmatrix} \bm b_V \\ \bm b_H \end{bmatrix} \in \Rd^N$ and $\bm W = \begin{bmatrix}
    \bm W_{V\gets V} & \bm W_{V\gets H} \\
    \bm W_{H\gets V} & \bm W_{H\gets H}
\end{bmatrix} \in \Rd^{N\times N}$. For visible neurons, $x_{t,n} \sim \operatorname{Poisson}(f_{t,n})$.

\subsubsection{Variational sampling schemes}
$\bullet$ \textbf{Homogeneous Poisson}
\begin{equation}
    f_{t,h} = \sigma(c_h).
\end{equation}
The variational parameter set is $\phi = \cbr{\bm c_H}$.\\
$\bullet$ \textbf{Inhomogeneous Poisson}
\begin{equation}
    f_{t,h} = \sigma(c_{t,h}).
\end{equation}
The variational parameter set is $\phi = \cbr{\bm C_{T\times H}}$.\\
$\bullet$ \textbf{Forward (F)}
\begin{equation}
    f_{t,h} = \sigma\rbr{c_h + \sum_{v'=1}^{V} a_{h\gets v'}\cdot \rbr{\sum_{l=1}^L x_{t-l,v'}\ \psi_l}}.
\end{equation}
The variational parameter set is $\phi = \cbr{\bm c_H,\bm A}$, where $\bm A = \begin{bmatrix}
    \bm O_{V\gets V} & \bm O_{V\gets H} \\
    \bm A_{H\gets V} & \bm O_{H\gets H}
\end{bmatrix} \in \Rd^{N\times N}$.\\
$\bullet$ \textbf{Forward-self (FS)}
\begin{equation}
    f_{t,h} = \sigma\rbr{c_h + \sum_{v'=1}^{V} a_{h\gets v'}\cdot \rbr{\sum_{l=1}^L x_{t-l,v'}\ \psi_l} + \sum_{h'=1}^H a_{h\gets h'} \cdot \rbr{\sum_{l=1}^L z_{t-l,h'}\ \psi_l}}.
\end{equation}
The variational parameter set is $\phi = \cbr{\bm c_H, \bm A}$, where $\bm A = \begin{bmatrix}
    \bm O_{V\gets V} & \bm O_{V\gets H} \\
    \bm A_{H\gets V} & \bm A_{H\gets H}
\end{bmatrix} \in \Rd^{N\times N}$.\\
$\bullet$ \textbf{Forward-backward (FB)}
\begin{equation}
    f_{t,h} = \sigma\rbr{c_n + \sum_{v'=1}^{V} a_{h\gets v'}\cdot \rbr{\sum_{l=1}^L x_{t-l,v'}\ \psi_l} + \sum_{v'=1}^V a_{v'\gets h} \cdot \rbr{\sum_{l=1}^L x_{t+l,v'}\ \psi_l}}
\end{equation}
The variational parameter set is $\phi = \cbr{\bm c_H,\bm A}$, where $\bm A = \begin{bmatrix}
    \bm O_{V\gets V} & \bm A_{V\gets H} \\
    \bm A_{H\gets V} & \bm O_{H\gets H}
\end{bmatrix} \in \Rd^{N\times N}$.

\clearpage
\subsubsection{Hidden spike distributions}
\begin{table}[!ht]
    \centering
    \caption{Different distributions of hidden spike, used in the generative model and the variational model. For simplicity, we omit the subscript of $z$, $\tilde{\bm z} = (\tilde z_0, \dots, \tilde z_{M-1})$, and $f$ indexing the hidden neuron and time bin. }
    \begin{tabular}{cccc}
        \toprule
        distribution & sample & likelihood & can use pathwise \\
        \midrule
        Poisson (Pois) & $z\sim \operatorname{Poisson}(f)$ & $\Pd[z;f] = \frac{f^z \eu^{-z}}{z!}$
        & \xmark \\
        categorical (Cat) & $z \sim \operatorname{Cat}(\bm \pi(f))$ & $\Pd[z;f] = \bm\pi(f)_z$ & \xmark \\
        Gumbel-Softmax (GS) & $\tilde{\bm z}_{t,h} \sim \operatorname{GS}(\bm\pi(f_{t,h});\tau)$ & Eq.~\ref{eq:GS_likelihood} bellow & \cmark \\
        exponential (Exp) & $z \sim \operatorname{Exp}\rbr{\frac{1}{f}}$ & $\Pd[z;f] = \frac{1}{f}\exp\rbr{-fz}$ & \cmark \\
        Rayleigh (Ray) & $z \sim \operatorname{Ray}\rbr{\sqrt{\frac{2}{\uppi}}f}$ & $\Pd[z;f] = \frac{\uppi z}{2f^2}\exp\rbr{-\frac{\uppi z^2}{4f^2}}$ & \cmark \\
        Half-normal (HN) & $z\sim \operatorname{HN}\rbr{\sqrt{\frac{\uppi}{2}}f}$ & $\Pd[z;f] = \frac{2}{\uppi f}\exp\rbr{-\frac{z^2}{\uppi f^2}}$ & \cmark \\
        \bottomrule
    \end{tabular}
    \label{tab:distributions}
\end{table}

In the above table, we used the function $\bm \pi$ to truncate a Poisson distribution to a categorical distribution,
\begin{equation}
    \bm\pi(f) = \scalebox{0.9}{$\displaystyle\rbr{1-\sum_{m=1}^{M-1} \frac{f^m \eu^{f}}{m!},\frac{f^1\eu^{f}}{1!},\dots,\frac{f^{M-1}\eu^{f}}{(M-1)!}}$}.
\end{equation}
The GS likelihood is
\begin{equation}\label{eq:GS_likelihood}
    \Pd[\tilde{\bm z};f] = \Gamma(M)\tau^{M-1} \rbr{\sum_{m=0}^{m-1} \frac{\bm\pi(f)_m}{\tilde z_m^\tau}} \prod_{m=0}^{M-1} \frac{\bm\pi(f)_m}{\tilde z_m^{\tau + 1}}
\end{equation}

\clearpage
\subsection{Supplementary Figures}\label{appendix:supplementary_figures}
\subsubsection{Synthetic dataset}
\begin{figure}[!ht]
    \centering
    \includegraphics[width=\linewidth]{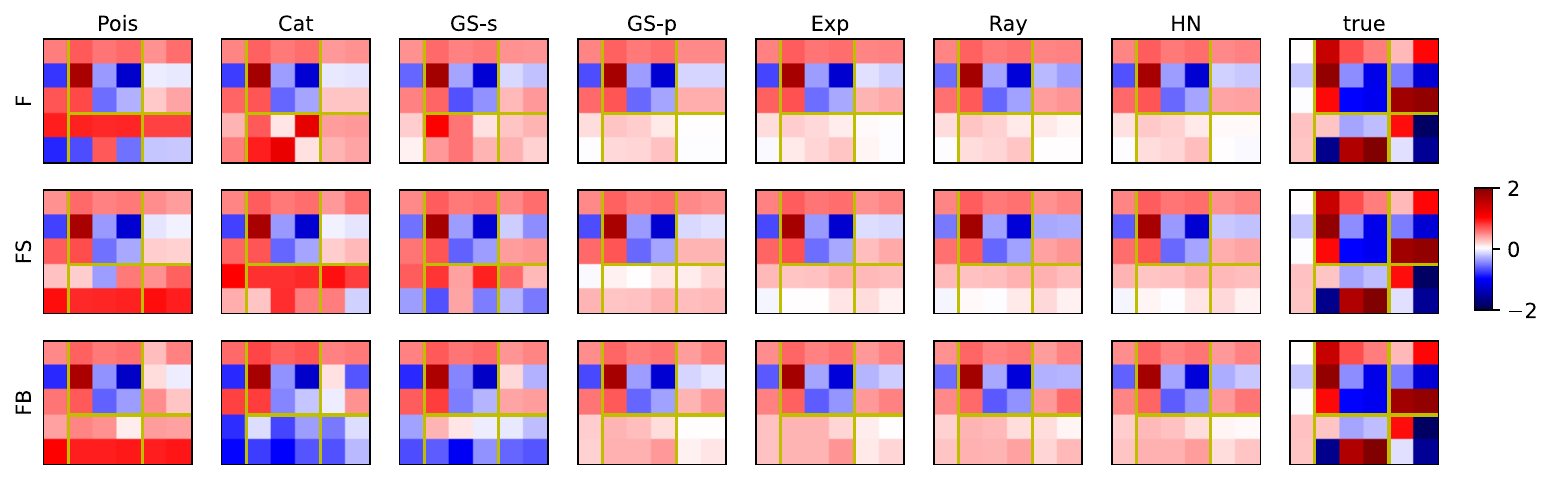}
    \caption{The learned weight matrix and bias vector compared with the true of all method combinations on one trial of the synthetic dataset.}
    \label{fig:synthetic_weight}
\end{figure}

\clearpage
\subsubsection{RGC dataset}
\begin{figure}[!ht]
    \centering
    \includegraphics[width=0.98\linewidth]{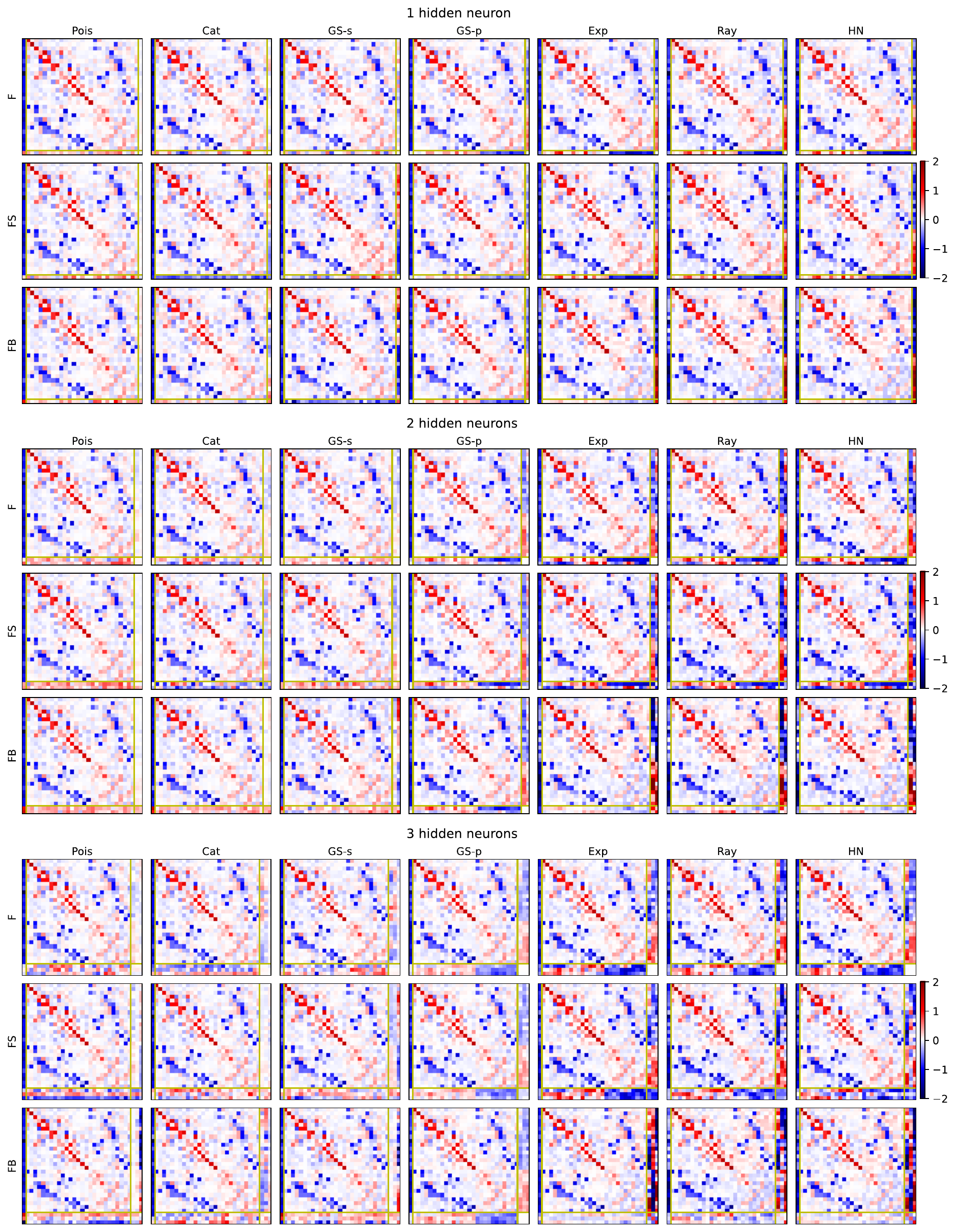}
    \vspace{-0.2in}
    \caption{The learned weight matrix and bias vector of all method combinations under different numbers of hidden neurons on the RGC dataset.}
    \label{fig:rgc_weight}
\end{figure}

\clearpage
\subsection{Point Process}\label{appendix:point_process}
\subsubsection{The generalized Hawkes process as the point process version of a GLM}
A generalized multivariate Hawkes process (GMHP) is a temporal point process (right-continuous) governed by the conditional intensity function
\begin{equation}
    \lambda_n^*(t) = \sigma\rbr{b_n + \sum_{t_n < t}w_{n\gets n_i}\cdot\psi(t-t_i)}
\end{equation}
where $n$ indexes for $N$ neurons. $(t_i, n_i)$ is the arrival time and neuron index of the $i$-th event in the history (spike) sequences ordered by arrival time $t_i$, i.e., $t_1 < t_2 < \dots < t$. $\lambda_n^*(t)$ is conditioned on the spike sequences before time $t$: $\cbr{(t_n,d_n)_{t_n<t}}$. $b_n\in\Rd$ is the background intensity of the $n$-th neuron, $w_{n\gets n'} \in\Rd$ is the connection weights from the $n'$-th neuron to the $n$-th neuron, and $\psi(\cdot)$ is the kernel function, which usually integrates to 1. $\sigma$ is a nonlinear function. Also, note that $\lambda_n^*(t)$ is a \textbf{left continuous} function. For preserving the causal relationship, we also need $\psi(t) = 0,\ \forall t\leqslant 0$. Let $\theta = \cbr{\bm b,\bm W} = \cbr{[b_1,\dots,b_N]^\Tu,(w_{n\gets n'})_{N\times N})}$ denote the parameter set we estimate. The data likelihood (probability density function) of a spike train in the format of a continuous timestamps sequence $\Xc = \cbr{(t_i,n_i)}_{i=1}^{I}$ within a specific observation period $[0,T]$ is
\begin{equation}\label{eq:continuous_ll}
    \Pd(\Xc;\theta) = \prod_{n=i}^I \lambda_{n_i}^*(t_i) \cdot \exp\sbr{-\sum_{n=1}^N \int_0^T \lambda_n^*(t)\ \du t}
\end{equation}

\paragraph{Relationship between conditional intensity and arrival interval.} For any temporal point process (not discretized), the number of events happening between $t$ and $t+\Delta t$ follows the Poisson distribution
\[
X\sim \mathcal{P}\rbr{\int_t^{t+\Delta t} \lambda^*(s)\ \du s} = \mathcal{P}(\varLambda(t+\tau) - \varLambda(t))
\]
where $\varLambda(t) \coloneqq \int_0^t \lambda^*(s)\ \du s$ is the compensator. If $\Delta t \to 0$,
\[
\int_t^{t+\Delta t} \lambda^*(s)\ \du s \approx \lambda^*(t) \Delta t
\]

The relationship between the intensity function $\lambda^*(t)$ and the next (starts from current time $t$) time interval $\tau$ distribution (PDF) is
\[
\begin{split}
    f(\tau) = & \underbrace{\lambda^*(t+\tau)}_{\text{an event happens at }t+\tau,\text{ density}} \underbrace{\frac{\rbr{\int_t^{t+\tau}\lambda^*(s)\ \du s}^0\eu^{-\int_t^{t+\tau}\lambda^*(s)\ \du s}}{0!}}_{\text{no event happens in the interval }(t,t+\tau),\text{ probability}} \\
    = & \lambda^*(t+\tau) \eu^{-\int_t^{t+\tau} \lambda^*(s)\ \du s}
\end{split}
\]
CDF is
\[
F(\tau) = 1 - \eu^{-\int_t^{t+\tau}\lambda^*(s)\ \du s}
\]
Therefore, modeling the intensity function is equivalent to modeling the arrival interval.

\paragraph{Sampling/Simulation.} We introduce two equivalent sampling algorithms: Ogata's thinning and the First-come-first-serve (FCFS). The probability density of the next event happening at time $t+\tau$ of neuron $n$ is
\begin{equation}
    f(\tau,n) = \prod_{n'=1}^N \eu^{-\int_t^{t+\tau} \lambda_{n'}^*(s)\ \du s}\ \lambda_n^*(t+\tau)
\end{equation}
which can be viewed as FCFS. Using Ogata's thinning method,
\begin{equation}
    f(\tau,n) = \eu^{-\int_t^{t+\tau}\sum_{n'=1}^N \lambda_{n'}^*(s)\ \du s}\ \sum_{n'=1}^N \lambda_{n'}^*(t+\tau)\ \frac{\lambda_n^*(t+\tau)}{\sum_{n'=1}^N \lambda_{n'}^*(t+\tau)}.
\end{equation}

\subsubsection{Discretization of the continuous point process}
In general, the integration in the complete-data likelihood does not have an analytical solution. A general solution is Monte Carlo integration, despite being suboptimal. A better choice is the quadrature rule, such as the Simpson rule. However, no matter what numerical techniques we use, we lose the continuous property of the process itself. Besides, the data structure that stores the spikes is usually not ideal. First, computing the log-likelihood of a point process requires sequential searching on lists of timestamps, which is more complicated than a direct matrix multiplication used in binned spike train data. Second, sampling hidden spike timestamps requires sorting, which is extremely time-consuming. Furthermore, the integration term in the log-likelihood function usually has no closed-form expression and hence still needs time discretization. Therefore, it is more convenient to discretize the point process data into binned spike count at the very beginning rather than deal with sequences of continuous timestamps. Therefore, it is more convenient to discretize the process rather than deal with continuous timestamps.

Now, we introduce the discretized version. denote the discretized spike, where $S = \frac{T}{\Delta t}$ is the total number of time bins. Then $x_{s,n}$ represents the number of spikes of neuron $n$ in the time interval $((s-1)\Delta t,s\Delta t)$. Then the discretized GMHP becomes the generalized linear model (GLM): $x_{s,d}\sim \mathcal{P}(\lambda_{s,n}^*\Delta t)$ where $\mathcal{P}$ denotes the Poisson distribution, and
\begin{equation}\label{eq:discrete_ll}
    \lambda_{s,n}^* = b_n + \sum_{x_{s',n'} > 0,s' < s} x_{s',n'} w_{n\gets n'}\psi((s-s')\Delta t) = b_n + \sum_{n'=1}^N w_{n\gets n'} \sum_{l=1}^L x_{s-l,n'}\psi_l,
\end{equation}
where $\lambda_{s,d}^*$ is still parameterized by $\theta$ and $\bm\psi^\Tu = [\psi(\Delta t),\dots,\psi(L\Delta t),]$. The complete-data likelihood is
\begin{equation}
    \Pd(\bm X;\theta) = \prod_{n=1}^N\prod_{s=1}^S \frac{(\lambda_{s,n}^*\Delta t)^{x_{s,n}} \eu^{-\lambda_{s,n}^*\Delta t}}{x_{s,n}!}.
\end{equation}

Now, we show that this likelihood converges to the form in the continuous case when the width of the time bin $\Delta t\to 0$. When taking the limit $\Delta t\to 0$, $x_{s,n}$ can either be $0$ or $1$. Therefore,
\begin{equation}\label{eq:limit}
    \begin{split}
        \lim_{\Delta t\to 0}\Pd(\bm X;\theta) = & \prod_{x_{s,n}=1} \lambda_{s,n}^*\Delta t \ \eu^{-\lambda_{s,n}^*\Delta t} \prod_{x_{s,n} = 0} \eu^{-\lambda_{s,n}^*\Delta t} \\
        = & (\Delta t)^N \prod_{i=1}^I \lambda_{n_i}^*(t_i;\theta)\  \exp\sbr{-\sum_{n=1}^N\int_0^T\lambda_n^*(t;\theta)\ \du t}
    \end{split}
\end{equation}
Since $(\Delta t)^N$ is a constant, we divide the above equation by $(\Delta t)^N$ and just get the probability density, which is identical to Eq.~\ref{eq:continuous_ll} in the continuous case. Therefore, solving the discretized problem is equivalent to solving the original continuous problem, as long as $\Delta t$ is small enough. And the solution converges to the continuous solution as $\Delta t\to 0$. In fact, the procedure shown in Eq.~\ref{eq:limit} is applicable to any point process and its discretized version.


\end{document}